%% file: neurips_2020.tex
\documentclass{article}

\usepackage[preprint,nonatbib]{neurips_2020}

\usepackage[utf8]{inputenc} 
\usepackage[T1]{fontenc}    
\usepackage{hyperref}       
\usepackage{url}            
\usepackage{booktabs}       
\usepackage{amsfonts}       
\usepackage{nicefrac}       
\usepackage{microtype}      

\usepackage{amsmath}        
\usepackage{graphicx}       
\usepackage{import}         
\usepackage[dvipsnames]{xcolor}   
\usepackage{comment}         
\usepackage{cite} 
\usepackage[utf8]{inputenc}
\usepackage{amsmath}

\title{ Exposing the Functionalities of Neurons for Gated Recurrent Unit Based Sequence-to-Sequence Model}

\author{
Yi-Ting Lee, Da-Yi Wu, Chih-Chun Yang, Shou-De Lin\\
 Department of Computer Science \\and Information Engineering, \\
National Taiwan University  \\
}
  
\begin{document}

\maketitle

\begin{abstract}
  The goal of this paper is to report certain scientific discoveries about a Seq2Seq model. It is known that analyzing the behavior of RNN-based models at the neuron level is considered a more challenging task than analyzing a DNN or CNN models due to their recursive mechanism in nature. This paper aims to provide neuron-level analysis to explain why a vanilla GRU-based Seq2Seq model without attention can achieve token-positioning. We found four different types of neurons: storing, counting, triggering, and outputting and further uncover the mechanism for these neurons to work together in order to produce the right token in the right position. 
\end{abstract}

\import{sections/}{section1-introduction.tex}

\import{sections/}{section2-experiment-setting.tex}
\import{sections/}{section3-methods.tex}
\import{sections/}{section4-results.tex}

\import{sections/}{section5-interaction.tex}
\import{sections/}{section6-related-works.tex}

\import{sections/}{section7-conclusion.tex}

\import{sections/}{Impact_statement}

\bibliography{neurips_2020.bib}
\bibliographystyle{plain}



\end{document}

%% file: sections/section1-introduction.tex
\definecolor{mygreen}{RGB}{76, 146, 58}
\definecolor{mypurple}{RGB}{148, 103, 189}
\definecolor{myorange}{RGB}{255, 127, 14}
\definecolor{myred}{RGB}{214, 39, 40}
\definecolor{myblue}{RGB}{31, 119, 180}

\section{Introduction}

\begin{figure*}
  \centering
  \includegraphics[width=0.9\textwidth]{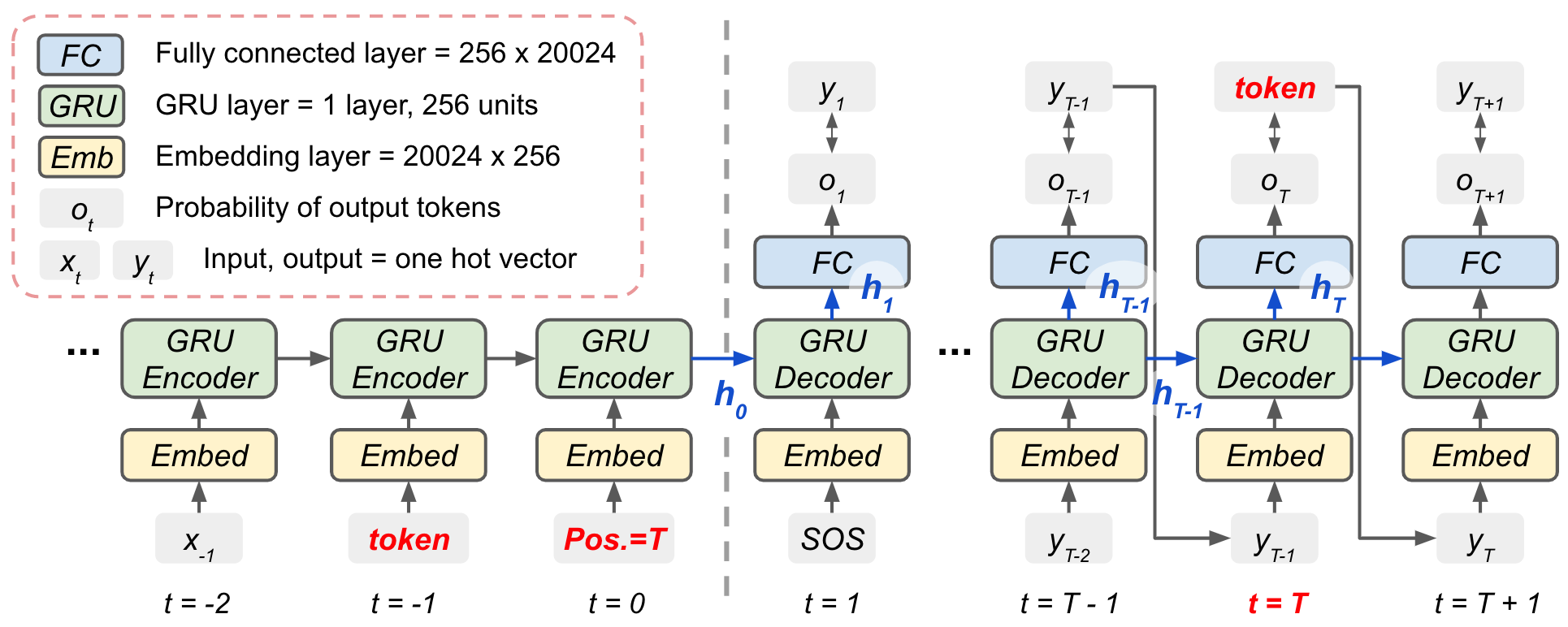}
  \caption{The Seq2Seq model combines classic encoder-decoder structure with one embedding layer and one fully connected layer without attention nor bi-directional structure.}
  \label{structure}
\end{figure*}

Sequence-to-sequence (Seq2Seq) \cite{sutskever2014sequence} models based on Recurrent Neural Networks \cite{mikolov2010recurrent} and its variant, such as Gated Recurrent Unit (GRU) \cite{cho2014learning} and Long Short-Term Memory (LSTM) cell \cite{hochreiter1997long}, has demonstrated high rate of success in areas such as question answering \cite{sukhbaatar2015end,dong2016language}, dialogue systems \cite{serban2016building}, machine translation \cite{sutskever2014sequence,luong2015multi}, and text generation \cite{wen2015stochastic, shang2015neural}. This paper focuses on a vanilla encoder-decoder based Seq2Seq model mostly following its original design in 2014 \cite{sutskever2014sequence}, as shown in Figure \ref{structure}, and the major difference is that we use GRU instead of LSTM. It is constructed by two recurrent blocks (encoder and decoder), an embedding layer to deal with inputs, and a fully connected layer to generate distributions of all tokens. In order to learn the output sequence distribution (target) given the input sequence (source), common training data are designed as input and output pairs: [$x_1,x_2,x_3$] $\rightarrow$ [$y_1,y_2,y_3,y_4$].

The inference process is described as follows: First the encoder encodes [$x_1,x_2,x_3$] to a dense vector $h_0$. Then at the first step of decoder, $h_0$ is transformed into $h_1$ after passing through the decoder unit once. $h_1$ then serves as input to the fully connected layer to produce the conditional probability of each single output token. Normally the token with highest probability becomes the first output token $\hat{y}_1$. $h_1$ shall be recursively passed down to the \emph{identical} decoder unit to generate conditional probability 
at subsequent time steps to produce tokens ($\hat{y}_2, \hat{y}_3, ...$) one by one.


Previous research \cite{shen2019controlling} has shown that it is possible to train a Seq2Seq model to output a token at a specific position with high accuracy. In fact, with two control signals, \emph{token} and \emph{position}, attached to the input sequence directly, they showed that a transformer model can learn the meaning of the control signals and output the right token at the right position with accuracy close to 100\%. 
Furthermore, it is also possible to train a vanilla Seq2Seq model (i.e. no attention or bidirectional structure is needed) like the one shown in Figure \ref{structure} to achieve the same goal. More precisely, training with sufficient number of examples containing token and position signals at the end of the input, the model can eventually learn that it has to output the token at the target position (e.g. third) as shown below:

\begin{center}
\textit{
$x_1$, $x_2$, $x_3$, {\color{myred}token}, {\color{mygreen}3} $\rightarrow$ $y_1$, $y_2$, {\color{myred}token}, $y_4$
}
\end{center}
 Note that a Seq2Seq auto-encoder\cite{ma2018autoencoder,xu2017variational} can be regarded as a realization of such token-positioning function since the model has to precisely outputs each token at the right position. 



We believe this finding is very interesting and potentially influential since the model in Figure \ref{structure} has no single component specifically designed for token positioning. It is exactly the same model used to perform machine translation as its original goal. 
Comparing with the state-of-the-art models that usually consist of much more complicated structure (e.g. transformer, GAN, etc.) or mechanism (e.g. attention, variational inference, etc.) to achieve the control of generation, we are quite surprised to learn that the simplest Seq2Seq model can already achieve fine-grained control of outputs given carefully designed training examples, which motivates us to focus on explaining the mechanism behind such capability. 

To begin with, we want to explain that generating a word at a target position is mathematically challenging for a vanilla Seq2Seq model as shown in Figure \ref{structure}. We start from Equation \ref{long-eq} that describes mathematically how the model produces output sequences during inference. 
The equation tells us that the target output $y_t$ is a fairly complicated function that contains $argmax$ among probabilities of all tokens in the vocabulary generate by the fully-connected layer FC, which depends on the GRU mapping of $h_{t-1}$ and  $y_{t-1}$, and further depends recursively on $h_{t-2}$ and  $y_{t-2}$, and so on. Eventually we can regard every output token as coming from a very complicated recursive function of the encoder output $h_0$. In order to output one specific token at position $t$, $h_0$ needs to make sure that after entering the same GRU unit recursively for $t-1$ times, the fully connected layer can assign the largest probability to the target token among tens of thousands of possible candidates. Also note that since the decoder shares the same FC and GRU function in every step, all the dynamic information needs to be stored in the hidden state $h_t$ in order to output different tokens at different positions.  

Therefore, each outputs ($\hat{y}_{1}$, $\hat{y}_2$, ..., $\hat{y}_T$) can be written mathematically as follows: 

\begin{equation}
    \begin{aligned}
    \hat{y}_1 &= \mathop{\arg\max}\limits_{y \in \text{vocab}} FC(h_1)
            = \mathop{\arg\max}\limits_{y \in \text{vocab}} FC \big(GRU(h_{0}, Emb(\hat{y}_{0}))\big) \\
    \hat{y}_2 &= \mathop{\arg\max}\limits_{y \in \text{vocab}} FC(h_2)
            = \mathop{\arg\max}\limits_{y \in \text{vocab}} FC \big(GRU({\color{myorange}h_{1}}, Emb({\color{myblue}\hat{y}_{1}}))\big) \\
            & = \mathop{\arg\max}\limits_{y \in \text{vocab}} FC \Big(GRU\big({\color{myorange}GRU(h_{0},Emb(\hat{y}_{0}))}\big),
            \quad  Emb({\color{myblue}\mathop{\arg\max}\limits_{y \in \text{vocab}} FC \big(GRU(h_{0}, Emb(\hat{y}_{0})))}\big)\Big) \\
    \hat{y}_t &=...        
    \end{aligned}
    \label{long-eq}
\end{equation}


In order to accurately position a token, this paper uncovers how a Seq2Seq model can perform the following functions: (1) The model not only needs to \emph{store} the token information after it appears in the encoder, but also confines its impact till the target outputt time. 
(2) In order to output at the right position, the Seq2Seq model requires a mechanism to \emph{count down}. (3) The \emph{storing} and \emph{counting} mechanisms need to interact in a certain way to ensure the fully connected layer assigns the largest probability to the target token. 

The contributions of this paper can be summarized as: (1) We find that neurons in a Seq2Seq model can perform four basic functions, \emph{storing, counting, triggering,} and \emph{outputting}. (2) We discover the mechanism behind counting down and token positioning based on the interaction among different types of neurons. (3) We propose a series of strategies to identify neurons of specific purpose in a recursive neural network, which can potentially be exploited for other types of neuron-based analysis. %

%% file: sections/section2-experiment-setting.tex
\section{Model and Training Details}

 We first describe the model and training details for achieving token-positioning. Terminology-wise, $t$ indicates the time step of a Seq2Seq model where $t=0$ indicates the end of encoding. Thus, negative $t$ corresponds to the encoding stage and positive $t$ is the decoding stage. We use $T$ to denote the target position for the assigned token.


\textbf{Dataset} We select two public datasets, the $Gutenberg$ dataset \cite{lahiri:2014:SRW} and Lyric dataset crawled from $Lyrics\text{ }Freak$, to generate the training data, and extract pairs of consecutive lines as the input and output sequences. We randomly select one token from the output sequence as our assigned token and concatenate the token and its position to the end of the input sequence. 
Note that we do not put any boundary signal to separate these control signals with the original input sequence, thus the model treats them just like it does to all other inputting tokens. The vocabulary size of our model is 20000. 
More details are shown in Appendix A.

\textbf{Model and Training Details} We choose a GRU-based encoder-decoder Seq2Seq model as described in Figure \ref{structure} (the parameters are also shown). Following Occum’s Razor policy, here we focus on the simplest possible architecture that can achieve such goal without complicating the analysis. 
It is important to point out that this vanilla model uses an identical bridge between encoder and decoder. In addition, the common perplexity loss of every token $= \exp(- \sum_{t=1}^{L+1} \log Pr(y_{t} | \hat{y}_{t-1}, h_{t-1}))$ is used as the training objective, where $y_t$ represents the target output, $\hat{y}_t$ as the model output, and $h_t$ as the hidden state. Note that during training we did not particularly assign loss to the token to be positioned, so the model is not directly guided to align the two control signals with the outputs. During inference, the model simply outputs the token with maximum probability, $\hat{y}_{t} = \mathop{\arg\max}\limits_{y \in \text{vocab}}Pr(y|\hat{y}_{t-1},h_{t-1})$, without adopting
a more complicated algorithm such as beam search. Besides, we use Adam as our optimizer and set early stop if the validation loss stops decreasing for few epochs. 
We focus the evaluation on whether the target token appears in the target position, and this token positioning accuracy can reach 98.5\% and 98.1\% respectively on the two datasets with random initialization multiple times. For generality, we substitute the GRU with LSTM models, and the average accuracy can also achieves 97.5\%. Finally, it is also possible to assign multiple control signals to control multiple tokens at the same time, and the accuracy achieves a respectful value of 95.2\%. It is surprising to us that by simply modifying the training dataset, a naive Seq2Seq model is capable of deciphering the message provided implicitly as Token and Position.

%% file: sections/section3-methods.tex
\section{Identifying Neurons of Specific Functions}

This section describes a general technique that allows us to identify and verify the critical neurons for certain basic functions. The process contains three stages: candidates generation, filtering, and verification by manipulating the neuron values. 


\subsection{Hypothesis formulation and candidate neurons generation}
We start from formulating certain hypothesis about the functions of neurons to be verified.
The first step to verify the hypothesis relies on checking whether it is possible to clearly separate instances that contain such function from those that do not. We realize such process by checking the accuracy of a classification task. The hypothesis is considered \emph{plausible} when the accuracy is higher than 95\%, or rejected otherwise. Once a hypothesis becomes plausible, we then perform feature-selection to pick a subset of dominate neurons as the responsible ones and move on to the next stage. 
A real example goes as follows. First a hypothesis is initiated: \emph{there is a subset of neurons in $h_t$ that stores the target token information}. Then we label all $h_t$ with a given token as positive and other instances NOT associated with this token as negative. Then we train a classifier using $h_t$ as features to examine whether it is possible to separate positive and negative sets. It turns out that we can reach training accuracy higher than 95\%, thus this hypothesis is considered plausible and we move on to select a subset of critical features.
In practice we choose linear support vector machine (SVM) as our basic classifier, and recursive feature elimination (RFE) as the feature selection mechanism.
In addition, at this stage we suggest less-rigorous selection to avoid missing candidate neurons, because the next two stages allow us to further confirm the hypothesis and remove false-positive neurons. 

\subsection{Filtering}

Neurons selected from the candidate generation stage only implies their patterns can be distinguished by an \emph{independently} trained classifier. However, it does not guarantee the original Seq2Seq model really utilizes them for specific purpose. Here we propose to use an internal mechanism to verify whether those neurons are critical to a specific output. We apply the method of integrated gradient \cite{sundararajan2017axiomatic} to quantify the importance of a neuron to the output and then filter out the less-relevant neurons. In integrated gradient, suppose we have an output function $F$, the integrated gradient score along with the i-th neuron for an input $x$ and baseline $x'$ is defined as follows:
\[\text{score}_i(x) = (x_i - x_i^{'}) \times \int_{\alpha = 0}^1 \frac{\partial F(x_i^{'} + \alpha (x_i - x'_i))}{\partial x_i} d \alpha\]
For instance, to measure the importance of each neuron in $h_t$ in terms of outputting token A at position T, we set $F(x)$ as the probability to output token A at t = T, and x as $h_t$. We then use the score to filter out the less-important neurons found in the \emph{candidate generation} step.



\subsection{Verification by manipulating the neuron values}
After the previous two stages, we have found a set of neurons that are influential to the output and \emph{potentially responsible} for a specific function. Here we propose to further verify them by replacing the value of these neurons with some carefully designed values and check if the behavior changes as expected. For example, we can replace the values of counting down neurons at $t$ with the corresponding values at 
$t - 1$, and check whether the model takes one more step to output the token. If it turns out that such replacement can indeed control the output behavior of the model, then we are confident to confirm the original hypothesis about the functionality of the neurons.  

%% file: sections/section4-results.tex
\section{List of Findings}\label{section4}

Here we report several scientific findings to better understand the mechanism of a GRU-based Seq2Seq model. 
To analyze the functional neurons, we exploit the 3-stage strategies described in the previous section on 50 most common tokens (occupying 48\% of the data) and common positions from 1 to 13 (occupying 75\% of the data). The detailed reports are documented in Appendix C. Below we summarize the key findings. 


\textbf{Finding 1.} For each token, regardless of its position, there is a small group of fixed neurons (15\% $\sim$ 25\%) jointly storing \emph{the token-specific information}. We call them \emph{storing neurons}. 
The storing neurons remain mostly unchanged from $t = -1$ (when the token signal appears in the input) to $T - 1$ (one step before the target position).


\begin{figure*}
    \centering
    \includegraphics[width = 0.9 \textwidth]{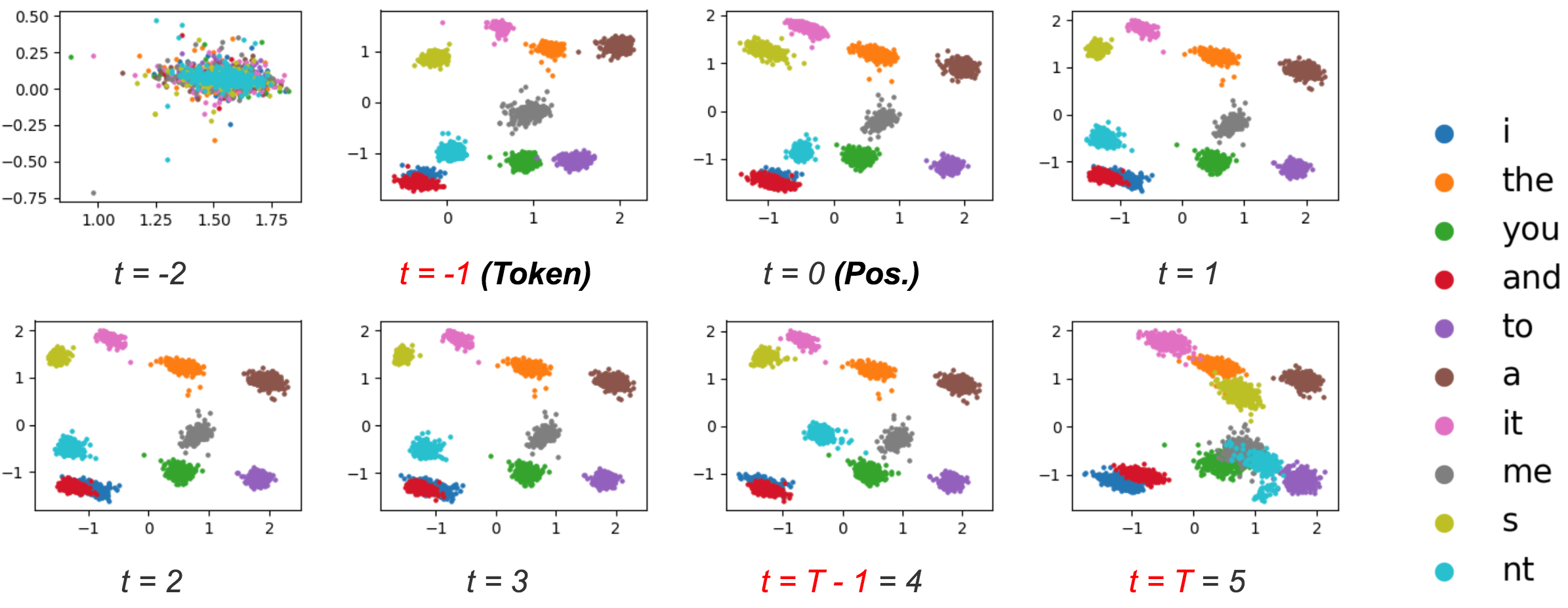}
    \caption{PCA plot of the storing neurons from ten most common tokens (500 samples each). From $t = -1$ to $T - 1$, the value of storing neurons hardly changes.}
    \label{S_store}
\end{figure*}

Our hypothesis is that before the target output time T, there exists a set of neurons who store the information about this token.
To verify this hypothesis, we first label the hidden states $h_t$ of samples that generate this token as positive and samples that generate other tokens as negative. The test is conducted on all 50 tokens and the result shows more than 95\% accuracy in classification.
Then the technique of feature selection is exploited to identify a subset of storing neurons in every $h_t$. We discover that the storing neurons (as well as their values) remain mostly unchanged across time stamps from $t=0$ to $t=T-1$. This implies that Seq2Seq model tends to use the same group of neurons with the same value to save a piece of information that needs to be stored for a long time. Figure \ref{S_store} shows the PCA projections of ten common tokens w.r.t. the progress of time to confirm the finding. Through visualization, we can show that the neurons for each token remains roughly the same until $t=T-1$. 
In the verification stage, we first verify that the storing neurons are necessary for model to output the correct token. Therefore, we replace the values of storing neurons at $t = T - 1$ with zero, and observe the accuracy drops significantly to $10.7\%$ and $20.0\%$, as shown in Table 1.
We also replace values of the storing neurons before $t = T - 1$ by their values in the previous time stamp, and observe the accuracy remains high (i.e. 87.9\% and 83.3\%), confirming that the values of the storing neurons remain roughly unchanged during the process.

\textbf{Finding 2.} Associated with each token, there is a small group of neurons (about 20\%) whose values keep increasing (or keep decreasing) with $t$ from $t = 0$ to $T - 1$. We call them \emph{counting neurons.} 

\begin{figure*}[h]
    \centering
    \includegraphics[width = 0.55\textwidth]{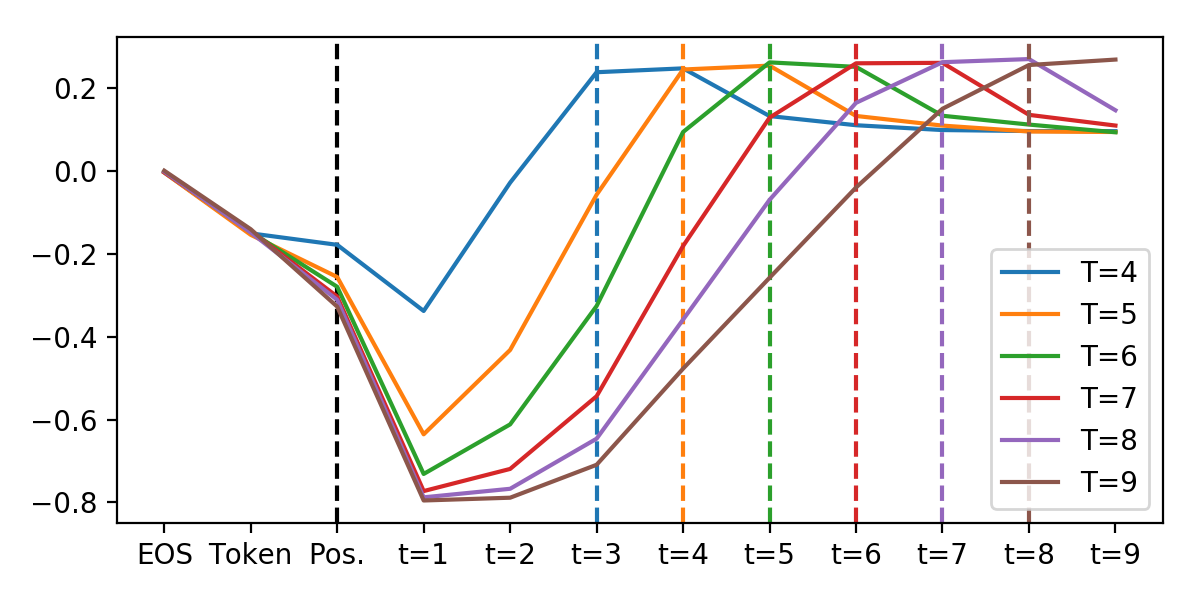}
    \caption{Average values of counting neurons with different assigned $T$. 
    Counting neurons with different $T$ scatter after seeing the \emph{position signal}, and they need different number of steps to reach the same end point.}
    \label{cd}
\end{figure*}

Here we assume that there are some neurons that shall function as \emph{counting down} toward the target position $T$. We first group all hidden states $h_t$ based on \emph{the number of steps before outputting (i.e. $T-t$)} as the labels. For example, the hidden state at step $t = 3$ given $T = 5$ and that of $t = 4$ given $T = 6$ share the same label $2$. Then we apply the classification and filtering strategy to identify a subset of neurons that are critical to the labels as counting neurons.
We then analyze the values of those counting neurons to understand how they can perform counting. The discovered behavior can be explained based on Figure \ref{cd}. When the position signal $T$ appears at the encoder at $t = 0$, the values of the counting neurons face significant drop. Furthermore, we find that the amount of dropping is positively correlated with the position $T$, meaning that the values move further toward negative when $T$ is larger. Then, as time step moves forward, the values of counting neurons slowly increase toward positive values. When reaching a relative high value, they will trigger another set of neurons (to be discussed in the later findings) to output the target token. Given different dropping range and mostly-constant increasing ratio, the counting neurons reach the optimal values after variable steps to achieve the function of counting (i.e. larger $T$ requires more steps to reach top). We find this phenomenon very unique and interesting. In hindsight, this seems to be a simple and elegant way for a memory-less, neuron-network-based model to perform counting. Given a well-trained decoder, achieving constant ratio increase of neuron values is not hard. Furthermore, as will be discussed in Section 5, the mechanism of GRU allows the activated counting neurons to trigger another set of neurons for firing the right token in the next step.  
We also observe some neurons perform counting in the opposite manner (i.e. starting by jumping to a positive range and gradually decreasing the values), but the underlying process remains the same.
Finally we verify that counting neurons can really control the token position. We replace the values of counting neurons at $t$ with those of the previous stage ($t - k$) and keep the rest of neurons as is, and then verify there are 99\% chance the target token can be generated $k$ step later, as shown in the counting part in Table \ref{table:table_verification} given $k=1 $to $3$.  
\textbf{Finding 3.} For each token, there is a small group of neurons (15\% $\sim$ 25\%) whose values change dramatically from $t < T - 1$ to $t = T - 1$, and when the above happens, the target token will be output at $t=T$. We call them \emph{triggering neurons}. It can be used as a powerful indicator to tell the model to take action in the next step.

To identify this type of neurons, we first train a classifier to separate $h_{t<T-1}$ from $h_{t=T-1}$, and then use feature selection and IG to select the subset of important neurons.
To verify the triggering neurons, we use their values at $t = T - 1$ to replace the same neurons at $t = T - k$ (with $k \geq 2$), and check whether the model can indeed output the assigned token earlier. Table \ref{table:table_verification} shows high accuracy while using the values of triggering neurons at t = 5 to overwrite the values at t = 1, 2, 3, and 4 respectively when T = 6. In the next section, we will further discuss why triggering neurons are required in the process. 

\begin{table}[h!]
	\centering
	\caption{The verification accuracy of the manipulation mentioned in Section 4.1, 4.2 and 4.3. The first two columns replace storing neurons with zeros or their previous state. The middle three columns show whether the output delay for $k$ steps when we replace the values of counting neurons at $t$ with those of the previous stage ($t - k$). The right four columns show whether the model can output the assigned token earlier when we use the triggering neurons to replace the earlier hidden state ($t - k$.} 
	\label{table:table_verification}
	\begin{tabular}{c|cc|ccc|cccc}
		\toprule
		Neuron set & \multicolumn{2}{c|}{Storing} & \multicolumn{3}{c|}{Counting} & \multicolumn{4}{c}{Triggering}\\
		Accuracy (\%) & Zero & Previous & k = 1 & k = 2 & k = 3 & k = 1 & k = 2 & k = 3 & k = 4 \\
		\midrule
		Lyrics Freak & 10.7 & 87.9 & 99.9 & 99.9 & 99.3 & 99.8 & 96.4 & 91.2 & 84.1 \\
		Gutenberg & 20.0 & 83.3 & 99.5 & 99.3 & 98.9 & 94.2 & 93.0 & 91.4 & 90.4 \\
	\bottomrule
	\end{tabular}
	
\end{table}



\textbf{Finding 4.} For each token, there is a small group of neurons (approximately 20\%) that is activated at $t=T$, causing the fully connected layer to assign highest probability to this token. We call them \emph{outputting neurons}.


The Seq2Seq model typically output the token with the highest probability. This probability is accumulated by a set of $h_{t=T} W_{token=A}$, where $W_{A}$ is the learned and fixed weights connected to target token A. To identify the outputting neurons, one can simply select the neurons with the largest $h_{t=T} W_{A}$. We find that only a small set of neurons in $h_{t=T}$ contributes to the output, and their values tend to be close to -1 or +1 (note that the range of GRU hidden states is -1 to +1). 
The verification process is designed as follows. We zero-out the values of outputting neurons and leave the rest of neurons on $h_{t=T}$ unchanged before feeding into the fully connected layer. It is observed that the accuracy of token positioning dropped to close to 0.0\%. In comparison, we zero-out all neurons (about 80\%) other than the outputting neurons, and find the accuracy remains as high as 99.9\%, confirming the function of these neurons.

\begin{figure*}
  \centering
  \includegraphics[width=0.9\textwidth]{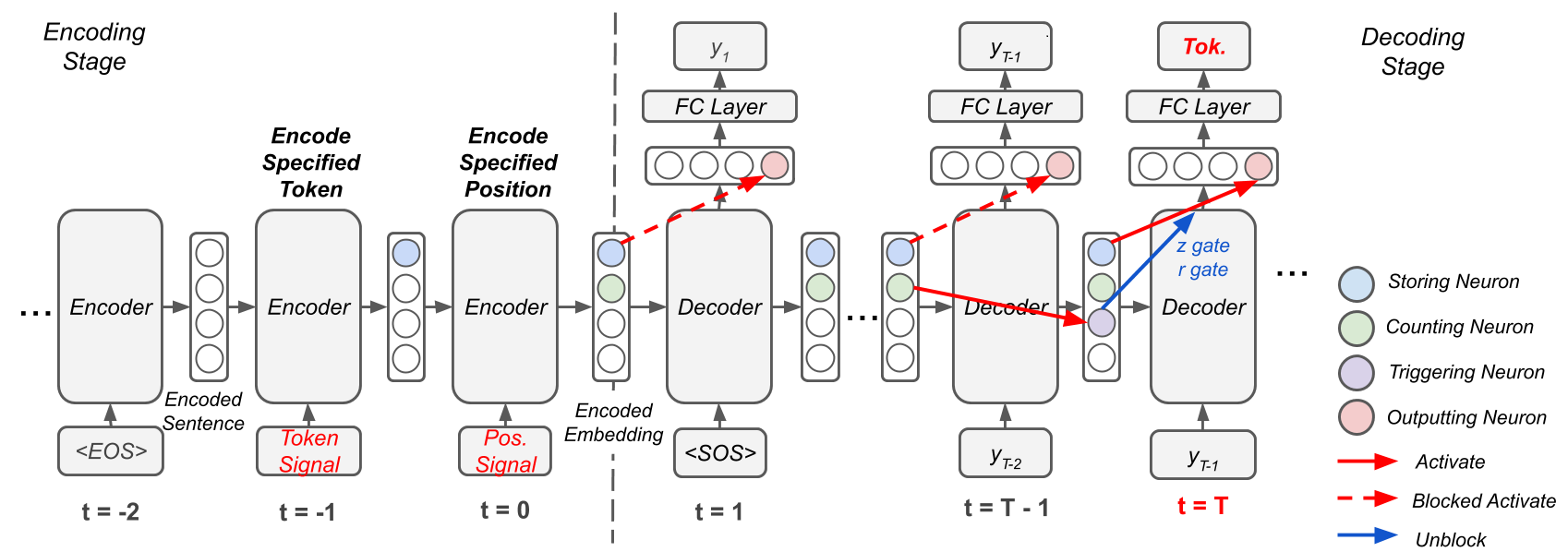}
  \caption{The interaction of four types of neurons for the token positioning task.}
  \label{flow_chart}
\end{figure*}

Figure \ref{flow_chart} is the conceptual graph to show how the neurons of different functions interact with each other to perform the token-positioning task. It can be summarized as below: 
(1) Storing neurons carry the token information (i.e. which to output) from $t = -1$ to $t = T - 1$. (2) The position information is carried by counting neurons starting from $t = 0$, and their values move piecewise linearly from $t = 0$ to $t = T - 1$. (3) Counting neurons reach the extreme values and activate the triggering neurons at $t=T-1$. (4) The triggering neurons and storing neurons then act together to fire outputting neurons at $t = T$, which then assigns target token with highest probability. 

In this process, there are several details that deserve special attention: (1) To output a target token at the right position, it implies that the model cannot output the token at other positions. Thus, the triggering neurons can also be regarded as playing a restraining role in the earlier stages. More details are shown in the next section. (2) Each token has its own four types of neuron sets but overlap may exist, as we discover that neurons can be multi-functional (e.g. play storing role before $t=T-1$ then shift to the outputting role at $t=T$).

%% file: sections/section5-interaction.tex
\section{Interaction of Neurons}



In this section, we want to discuss how the storing neurons and triggering neurons interact to activate the outputting neurons at $t = T$.
We start from a side-findings that, although the GRU decoder takes the previous state $h_{t-1}$ and the new embedded token $x_{t}$ (as shown in Figure \ref{structure}) as two inputs,  $x_{t}$'s role is negligible in token-positioning as we find that the positioning accuracy remains 98.0\% when all $x_t$ are replaced by random tokens. 
Consequently, we can modify the equation of GRU by eliminating the effects from external inputs $x$ and the constant bias terms to create a simplified version as Equation \ref{GRU'}: 

\begin{equation}
GRU\begin{cases}
    z_t = \sigma(W_{z}x_t + U_{z}h_{t-1} + b_z)  \\
    r_t = \sigma(W_{r}x_t + U_{r}h_{t-1} + b_r) \\
    \widetilde{h_t} = \text{tanh}(W_{h}x_t + U_{h}(r_t \odot h_{t-1}) + b_h)  \\ 
    h_t = z_t \odot h_{t-1} + (1 - z_t) \odot \widetilde{h}_t \\
    \end{cases} 
    \label{GRU}
\end{equation} 
\begin{equation}
GRU^{'}\begin{cases}
    z_t = \sigma(U_{z}h_{t-1})  \\
    r_t = \sigma(U_{r}h_{t-1}) \\
    \widetilde{h_t} = \text{tanh}(U_{h}(r_t \odot h_{t-1})) &  \\
    h_t = z_t \odot h_{t-1} + (1 - z_t) \odot \widetilde{h}_t\\
    \end{cases}
    \label{GRU'}
\end{equation}

where $z_t, r_t, h_t, \widetilde{h}_t$ are the \emph{update gate}, \emph{reset gate}, \emph{activation}, and \emph{candidate activation}, respectively. 
We can then view the operation inside the $\sigma(\cdot)$ function as a linear transformation of $h_{t-1}$ to $z_t, r_t$ or $h_t$. Therefore, $U_z, U_r, U_h$ can be regarded as the weights corresponding to the importance of neurons.
We already know that for $h_T[output]$ to generate the target token at step T, it needs to be significantly different from $h_{T-1}[output]$ (otherwise the target token will appear at time $T-1$), meaning that $z_t[output]$ has to be small and $\hat{h}_t[output]$ needs to be far away from zero according to the last part of Equation \ref{GRU'}.
Furthermore, in order to output the correct token, storing neurons need to be effective. It implies that $r_T[store]$ needs to be far from zero so that $h_{T-1}[store]$ can be influential. 
To sum up, only with large $r[store]$ and small $z[output]$, the token information can be passed from storing neurons to outputting neurons.

Then, we state our hypothesis as follows: it is the \emph{set of triggering neurons} that increases $r[store]$ and decreases $z[output]$ at the same time in order to prompt storing neurons to exert influence on outputting neurons.
We verify this hypothesis by conducting the following experiments. We mask out all neurons \emph{except triggering neurons} in $h_{t=T-1}$ to examine its impact $U_zh_{T-1}[output]$, and found that the values becomes negative, which causes $z_t[output] = \sigma({U_zh_{T-1}})[output]$ close to zero. We conducted similar experiments to verify that the triggering neurons can indeed increase $r[store]$.
Take the token "i" as an example, when triggering neurons are masked, the average of $z[output]$ increases from 0.067 to 0.410, and $r[store]$ drops from 0.860 to 0.594. It shows that although triggering neurons only occupy a small number of neurons, they can play most of the roles. More details are shown in Appendix D.


\begin{figure}[ht]
    \centering
    \includegraphics[width=1.0\textwidth]{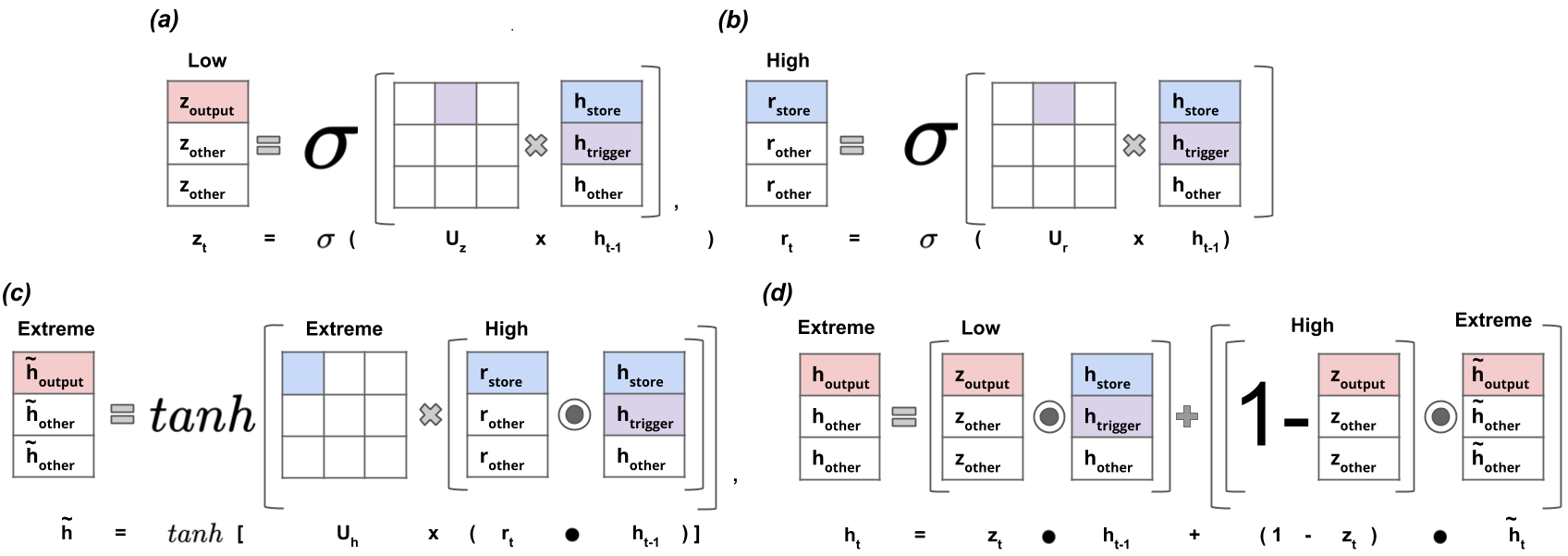}
    \caption{Visualization of neurons interaction. (a) Multiplying $U_z$ and triggering neuron decreases $z_{output}$. (b) Multiplying $U_r$ and triggering neuron increases $r_{store}$. (c) Large $r_{store}$ causes $h_{store}$ to become influential. (d) Small $z_{output}$ let $h_t[output]$ be updated by $\hat{h}$}
    \label{interaction}
\end{figure}

%% file: sections/section6-related-works.tex
\section{Related Works}

\textbf{Controlling Seq2Seq Model.}
	Previous works have shown that a Seq2Seq model can be controlled by directly manipulating the values in the hidden state or attaching additional signal to the encoded embedding. In \cite{Identify_Control_NMT}, it shows examples of changing the values of neurons responsible for tense and gender to control the outputs in machine translation.  \cite{weiss2018practical} proposes to control the attributes of generated sequences such as mood and tense by concatenating the attribute embedding to the encoded input sequence.
	Different from the previous works that tries to manipulate the model to achieve the controlling, this paper pays more attention to explain why the controlling can be achieved.  







\textbf{Token-Positioning.}
In \cite{shen2019controlling}, it reveals a wide variety of application involves the token-positioning task. For example, it is used to demo a acrostic poem system generating hidden token pattern through controlling the outputting words at certain positions. It can be used to produce rhyming lyrics by assigning tokens that rhymes at the end of the sequence.   Additionally, it can control the length of a sentence by assigning the token "EOS" at the target position. Seq2Seq auto-encoder\cite{ma2018autoencoder,xu2017variational} can 
also be regarded as a realization of positioning on multiple tokens since every token has an assigned position. 



\textbf{Understanding Recurrent Network Based on Neuron Analysis.}
There have been several works perform neuron-level analysis inside RNN for various purposes such as sentiment analysis \cite{RNN_Sentiment}, lexical and compositional storage \cite{qian-etal-2016-analyzing}, tense analysis \cite{Identify_Control_NMT}, and sentence structure analysis \cite{lakretz-etal-2019-emergence}.
One of the most commonly used strategy utilizes visualization to observe the dynamics of hidden states for each time-step \cite{Visualize_RNN}. 
 In \cite{weiss2018practical}, it shows that counting behavior can be achieved much easier in LSTM, while in other variants such as GRU, the dynamics of most neuron values seem to be irregular. Our findings can be considered as offering an explanation how GRU can perform counting even though each individual neuron's value does not seem to have strong correlation with the counts. 
 There have been some works utilizing diagnostic classifier \cite{giulianelli-etal-2018-hood},  Pearson Correlation \cite{Identify_Control_NMT}, and ablation test to probe the functions of neurons, which shares similar spirits to the first strategy we used (the only difference is that we further select a subset of dominant features). Nevertheless, as we have pointed out, simply performing diagnostic classification or correlation analysis cannot directly confirm the hypothesis, and the third stage of our solution (verification by manipulating neuron values) is the key factor to verify the functions of neurons. 
 Besides, \cite{Identify_Control_NMT} proposed to directly manipulate the neuron values to ameliorate the model performance, and similar strategy has been adopted by us to verify the functionality of neurons. Nevertheless, we believe the proposed 3-stage strategy can be a general mechanism that researchers can adopt to confirm the hypothesis of functions of neurons.

%% file: sections/section7-conclusion.tex
\section{Conclusion}
To perform fine-grained control on natural text generation, usually one would anticipate a more complicated structure such as attention-based or memory-based \cite{shen2013general,weston2014memory,sukhbaatar2015end} networks. What originally surprised us is that the vanilla Seq2Seq model can still achieve token positioning simply by learning from sufficient amount of training examples with control signals. The results from this paper show that a recurrent network with encoder-decoder structure could be much more powerful than one originally expected. Through the process of training, the neurons gradually developed their own capabilities to perform different functions such as storing, triggering, counting, etc. In the next stage, we will focus on uncovering the mystery behind the training process to learn how such dynamic neuron behavior can be trained with given samples.  




%% file: sections/Impact_statement.tex
\newpage
\vskip 0.075in%
{\large\bf Broader Impact}%
\vspace{1ex}%

This paper is NOT about proposing an ML model to achieve a certain goal nor reveals some methods to make a model more understandable to humans. This paper tries to answer two questions: 1. How to perform neuron-level analysis on a recurrent Seq2Seq model and 2. What did we discover after conducting such analysis?
Our findings include different functionality of neurons as well as how they interact to achieve a relatively challenging task given the underlying design of a Seq2Seq model. Among them, we believe the most exciting part is the discovery of the counting-down mechanism. Counting is not trivial for a memory-less model like the one we studied. We discover that the model achieves this by initializing the counting neurons with extreme values (e.g. -0.9), and then gradually add constant values  until reaching the top to trigger the next action. The amount of counting can be adjusted by assigning different level of initialization.

We hope this paper sends some messages to the community:
1. The first message is that more efforts is needed to investigated in terms of understanding the underlying mechanism of recurrent encoder-decoder networks. Since its first appearance in 2014, neural-based encoder-decoder based model has evolved so rapidly that almost in every year's top-tier conferences fancier models are presented, each of them more sophisticated yet more powerful than its predecessors. Though it is tempting and rewarding to design new and more complex models to pursue higher benchmark, here we hope to send a message that more efforts should be spent on trying to understand why they are more powerful. From 2014 till now, there have been thousands of papers published using (or improving) Seq2Seq models, but, based on our related work study, only dozens of the papers emphasize on analyzing and understanding the underlying behavior.  

2. The second signal we would like to send is that an encoder-decoder based Seq2Seq model based on GRU (or other RNN cells) might be more powerful than originally expected. We have revealed the token positioning capability, but based on our current analysis advanced control (e.g. control the rhyme or POS of each token) is also possible. 

3. Finally, this paper mainly focus on explaining the inference process. It remains a mystery to us how the model learns from training signal to develop different functionality. We believe understanding the training process of neurons might bring deeper insights to not just AI but also brain theory communities.